%% file: acl2023.tex
\pdfoutput=1
\documentclass[11pt]{article}

\usepackage[]{ACL2023}

\usepackage{times}
\usepackage{latexsym}

\usepackage[T1]{fontenc}
\usepackage[utf8]{inputenc}

\usepackage{microtype}

\usepackage{inconsolata}
\usepackage{xcolor}
\usepackage{amsmath,amsfonts,amssymb}
\usepackage{wasysym}  % empty, half, solid circles
\usepackage{makecell}
\usepackage{multirow}
\usepackage{multicol}
\usepackage{graphicx}
\usepackage{paralist}
\usepackage{url}
\usepackage{xspace}
\usepackage{amsfonts,amsmath}
\usepackage{booktabs}
\usepackage{setspace}
\usepackage{multirow}
\usepackage{multicol}
\usepackage{subfigure}
\usepackage[misc]{ifsym}
\usepackage{hyperref}
\usepackage{amssymb}% http://ctan.org/pkg/amssymb
\usepackage{pifont}% http://ctan.org/pkg/pifont
\newcommand{\cmark}{\ding{52}}%
\newcommand{\xmark}{\ding{56}}%
\usepackage{color, colortbl}
\usepackage{soul}
\usepackage{balance}

\definecolor{mydarkblue}{rgb}{0,0.08,0.45}
\definecolor{mydarkgreen}{HTML}{32612D}
\definecolor{myblue}{HTML}{3b75c3}
\definecolor{myred}{HTML}{E33222}
\definecolor{mygreen}{HTML}{438773}
\definecolor{mymaroon}{RGB}{142,27,19}
\definecolor{maroon}{HTML}{800000}
\definecolor{mycite}{cmyk}{0.55,1,0,0.15}
\definecolor{codeblue}{rgb}{0.25,0.5,0.5}
\definecolor{codekw}{rgb}{0.85, 0.18, 0.50}
\definecolor{codegreen}{rgb}{0,0.6,0}
\definecolor{codegray}{rgb}{0.5,0.5,0.5}
\definecolor{codepurple}{rgb}{0.58,0,0.82}
\usepackage{microtype}

\title{\vspace*{-0.5in}
{\normalsize \hfill Accepted to EMNLP 2024} \\
\vspace{0.35in}\textls[20]{Chain-of-Note: Enhancing Robustness in Retrieval-Augmented \\ Language Models}}

\author{Wenhao Yu, \ Hongming Zhang, \ Xiaoman Pan, \ Peixin Cao, \\ \textbf{Kaixin Ma, \ Jian Li, \ Hongwei Wang,  \ Dong Yu} \\
Tecent AI Lab \\
\tt wenhaowyu@global.tencent.com
}

\begin{document}
\maketitle

\begin{abstract}
\input{0-Abstract.tex}
\end{abstract}

\section{Introduction}
\input{1-Intro.tex}

\section{Proposed Method}
\input{4-Approach}

\section{Experiments}
\input{5-Experiments.tex}

\section{Related Work}
\input{6-Related.tex}

\section{Conclusion}
\input{7-Conclusion.tex}

\section{Limitations}
\input{8-Limitation.tex}

\bibliography{ref}
\bibliographystyle{acl_natbib}

\appendix

\clearpage
\section{Appendix}
\label{sec:appendix}
\input{9-Appendix}

% This is a section in the appendix.

\end{document}

%% file: 0-Abstract.tex
Retrieval-augmented language model (RALM) represents a significant advancement in mitigating  factual hallucination by leveraging external knowledge sources. 
However, the reliability of the retrieved information is not always guaranteed, and the retrieval of irrelevant data can mislead the response generation. 
Moreover, standard RALMs frequently neglect their intrinsic knowledge due to the interference from retrieved information. In instances where the retrieved information is irrelevant, RALMs should ideally utilize their intrinsic knowledge or, in the absence of both intrinsic and retrieved knowledge, opt to respond with "unknown" to avoid hallucination.
In this paper, we introduces \textsc{Chain-of-Note (CoN)}, a novel approach to improve robustness of RALMs in facing noisy, irrelevant documents and in handling unknown scenarios. The core idea of \textsc{CoN} is to generate sequential reading notes for each retrieved document, enabling a thorough evaluation of their relevance to the given question and integrating this information to formulate the final answer. 
Our experimental results show that GPT-4, when equipped with \textsc{CoN}, outperforms the \textsc{Chain-of-Thought} approach. 
Besides, we utilized GPT-4 to create 10K \textsc{CoN} data, subsequently trained on LLaMa-2 7B model.
Our experiments across four open-domain QA benchmarks show that fine-tuned RALMs equipped with \textsc{CoN} significantly outperform standard fine-tuned RALMs.

%% file: 1-Intro.tex
\begin{figure}[t]
    \centering
    \includegraphics[width=0.48\textwidth]{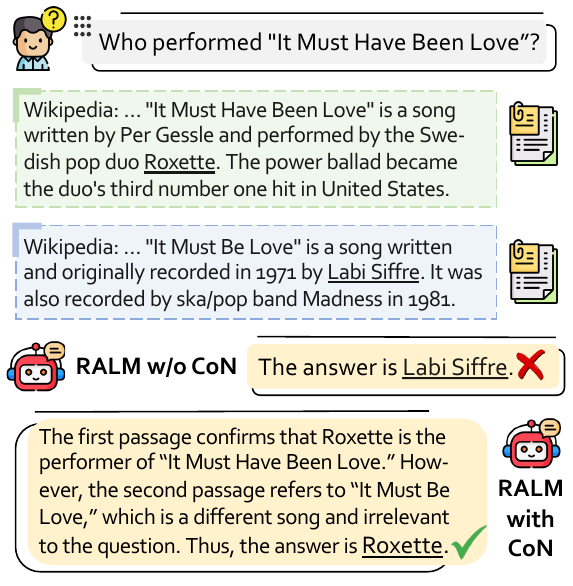}
    \vspace{-0.25in}
    \caption{Compared with the current RALMs, the core idea behind \textsc{Chain-of-Note (CoN)} is to generate sequential reading notes for the retrieved documents, ensuring a systematic assessment of their relevance to the input question before formulating a final response.}
    \label{fig:intro}
\end{figure}

Retrieval-augmented language models (RALMs) represent a novel framework that significantly advances large language models~\citep{touvron2023llama,openai2023gpt4} by addressing key limitations such as reducing factual hallucinations~\citep{ji2023survey,zhang2023siren}, injecting up-to-date knowledge in a plug-and-play manner~\citep{dhingra2022time,vu2023freshllms}, and enhancing domain-specific expertise~\citep{li2023chatgpt,qin2023chatgpt}. These enhancements primarily stem from integrating large language models (LLMs) with external knowledge sources~\citep{guu2020realm,lewis2020retrieval,borgeaud2022improving,shi2023replug}.
In a typical RALM setup, a query is first processed by a retriever that searches a vast evidence corpus for pertinent documents. A reader then examines these documents, extracting useful information and formulating the final output answer. 
% The potential benefit of the RALM framework is its ability to integrate relevant external knowledge, thereby enriching the LLMs' understanding of input text and generating answers based on this information. This is particularly beneficial when LLMs lack direct knowledge of a subject, allowing them to acquire and utilize relevant information in a plug-and-play manner~\citep{yu2022survey}.

% 3.2   equation? 
% separate out the ablation experiments setting?
% comparison against regular chain of thought? 
% model training/data sampling ratio experiments? 
% ChatGPT is then prompted with specific instructions tailored to the three distinct types of note generation. The quality of ChatGPT’s predictions is subsequently validated through human evaluations.

% Secondly, when building user-facing RALM applications, it's usually desirable to provide grounded answers to the user. However, existing RALMs mostly focus on the final answer prediction and often rely on additional post-processing modules to ground the response, which can be inefficient in practice. 
% The system should also provide fine-grained references along with its answer. 

However, there exist several issues with the current RALM framework.
First, there is no guarantee that the information retrieval (IR) system will always yield the most pertinent or trustworthy information. 
The retrieval of irrelevant data can lead to misguided responses~\citep{shi2023large,yoran2023making}, and potentially causing the model to overlook its inherent knowledge, even when it possesses adequate information to address the query~\citep{mallen2023not}.
Secondly, state-of-the-art LLMs often hallucinate when addressing fact-oriented questions, a deficiency that can be risky and may discourage users~\citep{ji2023survey,zhang2023siren}.
Ideally, an intelligent system should be capable of determining whether it has enough knowledge, both intrinsic and retrieved, to provide an accurate answer. In cases where knowledge is insufficient, the system should respond with ``unknown'' when the answer cannot be determined. Based on the shortcomings of the standard RALM system, in this paper, we aims to improve the robustness of RALMs, mainly focusing on two pivotal aspects: 

\textbf{(1) Noise Robustness:} The ability of a RALM to discern and disregard noisy information present in irrelevant retrieved documents, while appropriately leveraging its intrinsic knowledge.

\textbf{(2) Unknown Robustness:} The capacity of a RALM to acknowledge its limitations by responding with ``unknown'' when given a query it does not have the corresponding knowledge to answer, \textit{and} the relevant information is not found within the retrieved documents.

In this work, we introduce a novel framework named \textsc{Chain-of-Note (CoN)}, designed to enhance the robustness of RALMs. The cornerstone of \textsc{CoN} is to generate a series of reading notes for retrieved documents, enabling a comprehensive assessment of their relevance to the input query. This approach not only evaluates each document's pertinence but also pinpoints the most critical and reliable information therein. This process effectively filters out irrelevant or less credible content, leading to responses that are more precise and contextually relevant, as exemplified in Figure \ref{fig:intro}.
Besides, \textsc{CoN} enhances the capability of RALM to handle queries fall outside the scope of training data. In cases where the retrieved documents do not provide any relevant information, \textsc{CoN} can guide the model to acknowledge its limitations and respond with an ``unknown'' or provide possible explanation based on available data, enhancing reliability.

To validate the effectiveness of the \textsc{CoN} idea, we first conducted a comparison with \textsc{Chain-of-Thought (CoT)}~\cite{wei2022chain} using GPT-4 as the reader, showing \textsc{CoN} is more effective than \textsc{CoT} in retrieval-augmented scenarios. 
Next, we prompted GPT-4~\citep{openai2023gpt4} to generate a 10K training examples based on questions collected from NQ~\citep{kwiatkowski2019natural}, and subsequently trained on the LLaMa-2 7B, to valid the note-taking ability for smaller-sized models. Our evaluation of the RALM, integrated with \textsc{CoN} and compared to the standard RALM system, focused on three major aspects: (1) overall QA performance using DPR-retrieved documents, (2) noise robustness, assessed by introducing noisy information to the system, and (3) unknown robustness, evaluated through queries not covered in the LLaMa-2 pre-training data, i.e., real-time questions.
The evaluations were conducted on the NQ and three additional out-of-domain open-domain QA datasets, namely TriviaQA~\citep{joshi2017triviaqa}, WebQ~\citep{berant2013semantic}, and RealTimeQA~\citep{kasai2023realtime}. Our experiments show that \textsc{CoN} not only improves overall QA performance when employed with DPR-retrieved documents but also significantly enhances robustness in both noise and unknown aspects. This includes a +7.9 increase in accuracy (measured by the exact match score) with noisy retrieved documents, and a +10.5 increase in the rejection rate for real-time questions\footnote{We use real-time questions collected from RealTimeQA after May 2023, which was not trained by LLaMa-2.} that are beyond the pre-training knowledge scope.
% Through this paper, we aim to present a cohesive strategy for enhancing the capabilities of RALMs, ensuring their reliability amidst noisy information.

%% file: 4-Approach.tex
\begin{figure*}[t]
    \centering
    \includegraphics[width=0.99\textwidth]{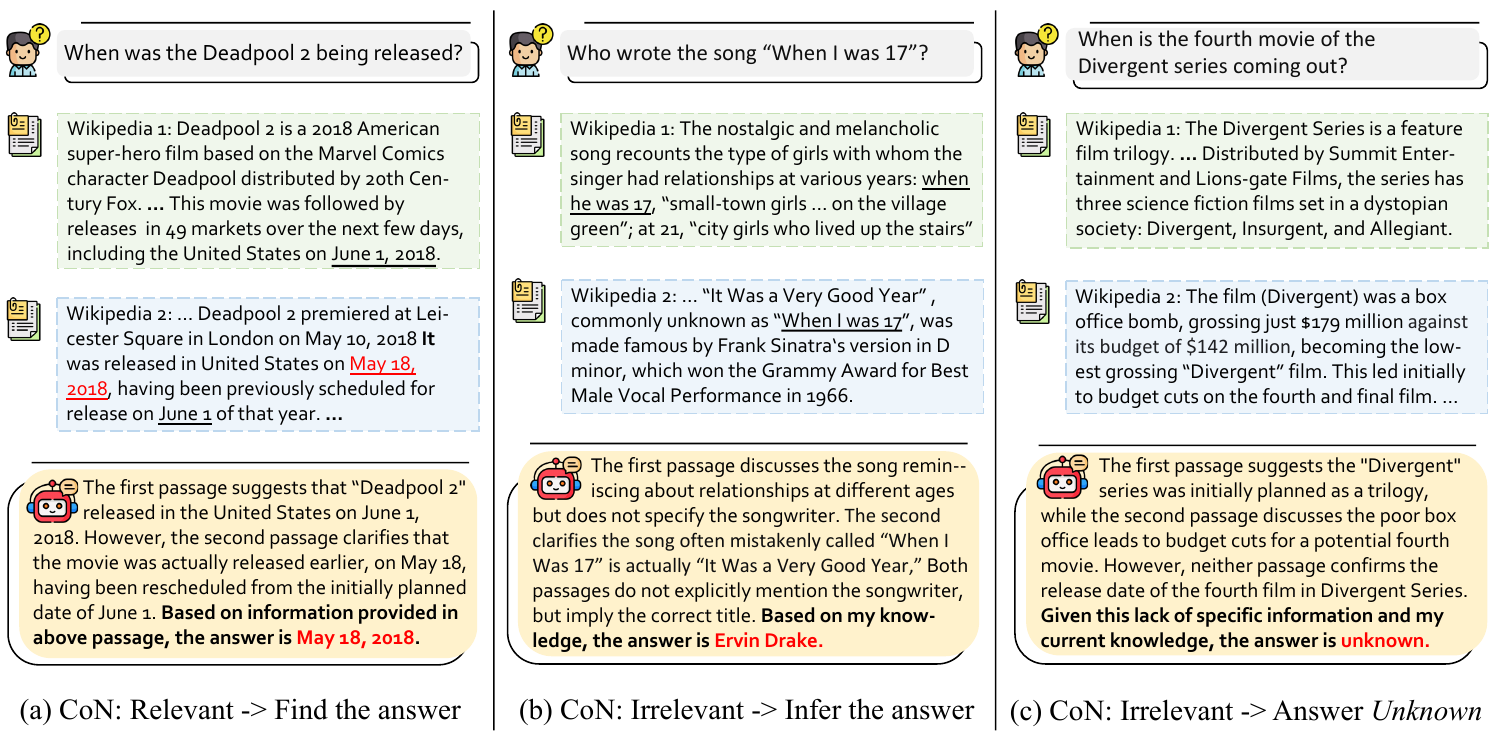}
    \vspace{-0.15in}
    \caption{Illustration of the \textsc{Chain-of-Note (CoN)} framework with three distinct types of reading notes. Type (a) depicts the scenario where the language model identifies a document that directly answers the query, leading to a final answer formulated from the retrieved information. Type (b) represents situations where the retrieved document, while not directly answering the query, provides contextual insights, enabling the language model to integrate this context with its inherent knowledge to deduce an answer. Type (c) illustrates instances where the language model encounters irrelevant documents and lacks the necessary knowledge to respond, resulting in an ``unknown'' answer. This figure exemplifies the CoN framework's capability to adaptively process information, balancing direct information retrieval, contextual inference, and the recognition of its knowledge boundaries. }
    \label{fig:method}
\end{figure*}

\subsection{Overview}

In this section, we introduce \textsc{Chain-of-Note}, an innovative advancement for retrieval-augmented language models (RALMs). Specifically, \textsc{CoN} framework generates sequential reading notes for the retrieved documents, which enables a systematic evaluation of the relevance and accuracy of information retrieved from external documents. By creating sequential reading notes, the model not only assesses the pertinence of each document to the query but also identifies the most critical and reliable pieces of information within these documents. This process helps in filtering out irrelevant or less trustworthy content, leading to more accurate and contextually relevant responses.

\subsection{Background of Existing RALMs}

RALMs signify a transformative development in language models, enhancing their output by incorporating external knowledge. These models operate by introducing an auxiliary variable, denoted as $d$, which represents retrieved documents. This inclusion allows them to consider a range of possible documents, thereby producing responses that are more informed and precise \citep{lazaridou2022internet,shi2023replug}.
The RALM models can be represented as \(p(y|x) = \sum_{i} p(y|d_i, x) p(d_i|x)\). Here, $x$ represents the input query, and $y$ signifies the model's generated response.
In practice, it is infeasible to compute the sum over all possible documents due to the vast number of potential sources. Consequently, the most common approach involves approximating the sum over $d$ using the $k$ highest ranked documents, and providing all these documents as part of the input. 
We assume, w.l.o.g., that these documents are $[d_1, \ldots, d_k]$, yielding $p(y|x) = \sum_{i=1}^k p(y|d_i, x) p(d_i|x)$. 

However, the existing RALMs suffer from several limitations: 

$\bullet$ Risk of Surface-Level Processing: When directly generating an answer, language models might rely on surface-level information without deep comprehension. Thus, they could easily overlook the nuances of question or documents, particularly in complex or indirect questions.

$\bullet$ Difficulty in Handling Contradictory Information: When faced with documents containing contradictory information, directly generating an answer becomes challenging. The model may struggle of these contradictions or to determine which piece of information is more credible or relevant.

$\bullet$ Reduced Transparency and Interpretability: Direct answer generation offers limited insight into how the model arrived at its conclusion. This lack of transparency makes it challenging for users to understand the basis of the model's conclusions.

$\bullet$ Overdependence on Retrieved Documents: Direct generation can lead to an overreliance on the content of the retrieved documents (i.e. tendency to extract information from retrieved documents~\citep{shi2023large}), ignoring the model's inherent knowledge base. This can be particularly limiting when the retrieved documents are noisy or out-of-date.

\subsection{The \textsc{Chain-of-Note} Framework}
\label{sec:method}

The \textsc{Chain-of-Note (CoN)} framework presents a solution to the challenges faced by retrieval-augmented language models (RALMs). This framework significantly enhances the ability of RALMs to critically assess retrieved documents through a structured note-taking process. Specifically, it involves generating concise and contextually relevant summaries or notes for each document. This method allows the model to systematically evaluate the relevance and accuracy of information drawn from external documents. By creating sequential reading notes, \textsc{CoN} not only assesses the pertinence of each document to the query but also pinpoints the most reliable information and resolves conflicting information. This approach effectively filters out irrelevant or less trustworthy content, leading to responses that are both more accurate and contextually relevant.

Given an input question $x$ and $k$ retrieved documents $[d_1, \cdots, d_k]$, the model aims to generate textual outputs comprising multiple segments $[y_{d_1}, \cdots, y_{d_k}, y]$. Here, $y_{d_i}$ signifies the tokens for the $i$-th segment, representing the reading note for the corresponding document $d_i$, as shown in Figure \ref{fig:method}. After generating individual reading notes, the model synthesizes the information to create a consolidated final response $y$. The implementation of the \textsc{Chain-of-Note (CoN)} involves three key steps: (1) designing the notes $y_{d_i}$, (2) collecting the data, and (3) training the model.

\subsubsection{\textsc{Chain-of-Note} Format Design}

The framework primarily constructs three types of reading notes, as shown in Figure \ref{fig:method} , based on the relevance of the retrieved documents to the input question: First, when a document directly answers the query, the model formulates the final response based on this relevant information, as shown in Figure \ref{fig:method}(a). Second, if the retrieved document does not directly answer the query but provides useful context, the model leverages this information along with its inherent knowledge to deduce an answer, as shown in Figure \ref{fig:method}(b).
Third, in cases where the retrieved documents are irrelevant, and the model lacks sufficient knowledge to answer, it defaults to responding with ``unknown", as shown in Figure \ref{fig:method}(c).
This nuanced approach mirrors human information processing, striking a balance between direct retrieval, inferential reasoning, and the acknowledgment of knowledge gaps.

\subsubsection{Data Collection and Model Training}

To equip the model with the ability to generate such reading notes, it's essential to gather appropriate training data. Manual annotation for each reading note is resource-intensive, so we employ a state-of-the-art language model -- GPT-4 -- to generate the notes data. This method is both cost-effective and enhances reproducibility.
We initiate this process by randomly sampling 10k questions from the NQ~\citep{kwiatkowski2019natural} training dataset. GPT-4 is then prompted with specific instructions and in-context examples to the three distinct types of note generation (detailed in Appendix \ref{sec:prompt}). The quality of GPT-4's predictions is subsequently assessed through human evaluations on a small subset of the data before proceeding to the entire set.
The NQ dataset is chosen as our primary dataset due to its diverse range of real user queries from search engines. However, to ensure the model's adaptability, we also test its performance on three additional open-domain datasets, including TriviaQA, WebQ, and RealTimeQA, showing its generalization capabilities to out-of-domain (OOD) data.

After collecting 10K training data from GPT-4, the next step involves using them to train a LLaMa-2 7B model~\cite{touvron2023llama}, to validate the feasibility of generating \textsc{Chain-of-Note (CoN)} outputs.
To do this, we concatenate the instruction, question and documents as a prompt and train the model to generate notes and answer in a standard supervised way.
Our in-house model learns to sequentially generate reading notes for each document to assess their relevance to the input query. Responses are generated based on the document's relevance, enhancing accuracy and reducing misinformation. If all documents are irrelevant, the model either relies on inherent knowledge for an answer or responds with ``unknown'' if the answer cannot be determined  accurately.

\subsubsection{Hybrid Training for Better Efficiency}

Generating \textsc{Chain-of-Note (CoN)} would increase inference cost, potentially hindering real-world usage. To address this, we experimented with a simple yet effective strategy for internalizing \textsc{CoN} reasoning, called \textit{Hybrid Training}.

Specifically, we allocate 50\% of the training time to the standard RALM, which involves directly generating answers without notes, and the other 50\% to RALM with CoN. This strategy allows the model to internalize intermediate reasoning steps during training. Additionally, we add two different prompt words before each category of data.

During the inference phase, we exclusively use the standard RALM prompt to guide the model, prompting it to output answers \textit{without} relying on explicit reading notes. This approach leverages the hidden states developed during training for implicit \textsc{CoN} reasoning. The model trained with the hybrid training strategy maintains the same inference time while achieving only slightly lower performance wit CoN. The results will be introduced in $\S$\ref{sec:hybrid}.

%% file: 5-Experiments.tex
\begin{table}
\centering
\setlength{\tabcolsep}{2mm}{
\scalebox{0.95}{\begin{tabular}{l|ccc}
\toprule
Datasets & Full size & IR Recall & Subset size \\
\midrule
NQ & \ 3,610 & 73.82 & 2,086 \\
TriviaQA & \ 7,993 & 89.95 & 7,074 \\
WebQ & \ 2,032 & 64.22 & 1,231 \\
\bottomrule
\end{tabular}}}
\vspace{-0.1in}
\caption{Dataset statistics. The recall evaluation is based on DPR retrieval on the full test set. }
\label{tab:stats}
\end{table}

\begin{table*}[t]
\centering
\setlength{\tabcolsep}{2.7mm}{
\scalebox{0.9}{\begin{tabular}{l|cc|cc|cc|cc}
\toprule
\multirow{2}{*}{Models} & \multicolumn{2}{c|}{NQ} & \multicolumn{2}{c|}{TriviaQA} & \multicolumn{2}{c|}{WebQ} & \multicolumn{2}{c}{Average} \\
& EM & F1 & EM & F1 & EM & F1 & EM & F1 \\ 
\midrule
\multicolumn{9}{l}{\textit{Backbone language model: LLaMa-2 7B}} \\
QA fine-tune w/o IR & 28.80 & 37.53 & 63.19 & 68.61 & 28.30 & 42.77 & 35.98 & 44.27 \\
SAIL~\cite{luo2023sail}* & 36.20 & 44.23 & 73.20 & 80.92 & 27.92 & 40.65 & 45.77 & 55.27 \\
\midrule
Retrieve-Read~\cite{shi2023replug} & 47.39 & 55.81 & 74.92 & 81.53 & 29.58 & 43.51 & 48.49 & 56.97 \\
\ \ + \textsc{Chain-of-Note} (ours) & 48.92 & 57.53 & 76.27 & 82.25 & 32.33 & 46.68 & 50.46 & 58.78  \\
& \small{\textbf{(+1.53)}} & \small{\textbf{(+1.72)}}& \small{\textbf{(+1.35)}}& \small{\textbf{(+0.72)}}& \small{\textbf{(+2.75)}}& \small{\textbf{(+3.17)}}& \small{\textbf{(+1.97)}}& \small{\textbf{(+1.81)}} \\
\midrule
\midrule
\multicolumn{9}{l}{\textit{Backbone language model: \texttt{GPT-4-1106}} $\dagger$} \\
QA prompt w/o IR & \multicolumn{2}{c|}{54.0} & \multicolumn{2}{c|}{74.2} & \multicolumn{2}{c|}{56.2} & \multicolumn{2}{c}{61.5} \\
% \midrule
Retrieve-Read~\cite{shi2023replug} & \multicolumn{2}{c|}{61.8} & \multicolumn{2}{c|}{70.6} & \multicolumn{2}{c|}{56.8} & \multicolumn{2}{c}{63.1} \\
\ \ + \textsc{Chain-of-Thought} & \multicolumn{2}{c|}{63.6} & \multicolumn{2}{c|}{71.2}  & \multicolumn{2}{c|}{58.4} & \multicolumn{2}{c}{64.4} \\
\ \ + \textsc{Chain-of-Note (ours)} & \multicolumn{2}{c|}{63.8} & \multicolumn{2}{c|}{74.6} & \multicolumn{2}{c|}{58.8} & \multicolumn{2}{c}{65.7} \\
& \multicolumn{2}{c|}{\small{\textbf{(+2.0)}}} & \multicolumn{2}{c|}{\small{\textbf{(+4.0)}}} & \multicolumn{2}{c|}{\small{\textbf{(+2.0)}}} & \multicolumn{2}{c}{\small{\textbf{(+2.6)}}}\\
\bottomrule
\end{tabular}}}
\vspace{-0.1in}
\caption{The RALM, when equipped with \textsc{Chain-of-Note (CoN)}, demonstrates a marginal improvement over the standard RALM in full test set evaluations. Significantly, it outperforms the standard RALM system in scenarios with noisy documents, suggesting that \textsc{CoN} can substantially enhance the model's noise robustness.
\\ * SAIL was designed for retrieval-augmented instruction tuning, and as such, may not be ideally factual QA.
\\ $\dagger$ Evaluating GPT-4 outputs with EM score is challenging; we opt for Accuracy, with reasons outlined in $\S$\ref{sec:metric}.
}
\label{tab:overall}
\end{table*}

\begin{table*}[t]
\centering
\setlength{\tabcolsep}{2.5mm}{
\scalebox{0.9}{\begin{tabular}{l|c|cc|cc|cc|cc}
\toprule
\multirow{2}{*}{Models} & Noise & \multicolumn{2}{c|}{NQ} & \multicolumn{2}{c|}{TriviaQA} & \multicolumn{2}{c|}{WebQ} & \multicolumn{2}{c}{Average} \\
& Ratio & EM & F1 & EM & F1 & EM & F1 & EM & F1 \\ 
\midrule
Retrieve-Read & \multirow{2}{*}{100\%} & 34.28 & 41.74 & 55.30 & 61.67 & 29.58 & 46.34 & 39.72 & 49.92 \\
\ \ + \textsc{Chain-of-Note} & & 41.83 & 49.58 & 64.30 & 70.00 & 36.85 & 53.07 & 47.66 & 57.55 \\
& & \small{\textbf{(+7.55)}} & \small{\textbf{(+7.84)}}& \small{\textbf{(+9.00)}}& \small{\textbf{(+8.33)}}& \small{\textbf{(+7.27)}}& \small{\textbf{(+6.73)}}& \small{\textbf{(+7.94)}}& \small{\textbf{(+7.63)}} \\
\midrule
Retrieve-Read & \multirow{2}{*}{80\%} & 54.28 & 61.03 & 73.83 & 80.02 & 35.46 & 52.70 & 54.52 & 64.58 \\
\ \ + \textsc{Chain-of-Note} & & 56.63 & 63.23 & 75.89 & 81.24 & 40.60 & 56.54 & 57.70 & 67.00 \\
& & \small{\textbf{(+2.35)}} & \small{\textbf{(+2.20)}}& \small{\textbf{(+2.06)}}& \small{\textbf{(+1.22)}}& \small{\textbf{(+5.14)}}& \small{\textbf{(+3.84)}}& \small{\textbf{(+3.18)}}& \small{\textbf{(+2.42)}} \\
\midrule
Retrieve-Read & \multirow{2}{*}{60\%} & 61.44 & 67.94 & 78.44 & 83.65 & 37.01 & 54.16 & 58.96 & 68.58 \\
\ \ + \textsc{Chain-of-Note} & & 63.43 & 69.33 & 78.79 & 84.07 & 41.26 & 56.91 & 61.16 & 70.10 \\
& & \small{\textbf{(+1.99)}} & \small{\textbf{(+1.39)}}& \small{\textbf{(+0.35)}}& \small{\textbf{(+0.42)}}& \small{\textbf{(+4.25)}}& \small{\textbf{(+2.75)}}& \small{\textbf{(+2.20)}}& \small{\textbf{(+1.52)}} \\
\midrule
Retrieve-Read & \multirow{2}{*}{40\%} & 64.62 & 71.12 & 80.56 & 86.76 & 38.40 & 55.60 & 61.19 & 71.16 \\
\ \ + \textsc{Chain-of-Note} & & 65.91 & 72.22 & 81.72 & 87.11 & 42.16 & 58.15 & 63.26 & 72.49 \\
& & \small{\textbf{(+1.29)}} & \small{\textbf{(+1.10)}}& \small{\textbf{(+1.16)}}& \small{\textbf{(+0.35)}}& \small{\textbf{(+3.76)}}& \small{\textbf{(+2.55)}}& \small{\textbf{(+2.07)}}& \small{\textbf{(+1.33)}} \\
\midrule
Retrieve-Read & \multirow{2}{*}{20\%} & 67.21 & 73.69 & 81.73 & 87.89 & 39.95 & 56.66 & 62.96 & 72.75 \\
\ \ + \textsc{Chain-of-Note} & & 70.00 & 76.08 & 82.86 & 88.24 & 44.36 & 60.13 & 65.74 & 74.82 \\
& & \small{\textbf{(+2.79)}} & \small{\textbf{(+2.39)}}& \small{\textbf{(+1.13)}}& \small{\textbf{(+0.35)}}& \small{\textbf{(+4.41)}}& \small{\textbf{(+3.47)}}& \small{\textbf{(+2.78)}}& \small{\textbf{(+2.07)}} \\
\midrule
Retrieve-Read & \multirow{2}{*}{0\%} & 69.23 & 75.57 & 83.34 & 89.44 & 42.24 & 58.59 & 64.93 & 74.53 \\
\ \ + \textsc{Chain-of-Note} & & 73.28 & 79.86 & 83.52 & 88.94 & 46.16 & 62.38 & 67.65 & 77.06 \\
& & \small{\textbf{(+4.05)}} & \small{\textbf{(+4.29)}}& \small{\textbf{(+0.18)}}& \small{\textbf{(-0.50)}}& \small{\textbf{(+3.92)}}& \small{\textbf{(+3.79)}}& \small{\textbf{(+2.72)}}& \small{\textbf{(+2.53)}} \\
\bottomrule
\end{tabular}}}
\vspace{-0.1in}
\caption{Evaluation on Noise Robustness. The backbone language model is LLaMa-2 7B. The \textsc{Chain-of-Note} framework shows superior performance compared to the standard RALM system, particularly notable at higher noise ratios.We explain how we synthesize data with different noise ratios under real-world scenarios in $\S$ \ref{sec:split}.}
\label{tab:noise}
\end{table*}

\subsection{Experimental Settings and Evaluations}

\subsubsection{Datasets and Splits}
\label{sec:split}
We conducted comprehensive experiments using three benchmark datasets in open-domain question answering (QA): NQ~\citep{kwiatkowski2019natural}, TriviaQA~\citep{joshi2017triviaqa}, and WebQ~\citep{berant2013semantic}, with further details provided in Appendix \ref{sec:dataset}. Additionally, we employed RealTimeQA~\citep{kasai2023realtime} as a special case to evaluate ``unknown'' robustness.

The evaluation was conducted based on two evaluations sets: \textbf{full set and subset evaluation.} 
Firstly, akin to traditional open-domain QA evaluation, we assessed the models using all questions from the test set to evaluate the \textbf{overall QA performance}. The documents were retrieved using DPR, and the top-$k$ documents were fed into the generator. We adhered to the same test splits for the open-domain QA setting as used by \citet{izacard2021leveraging,karpukhin2020dense}. For TriviaQA, evaluations from LLaMa-2~\citep{touvron2023llama} were conducted on the Wikipedia dev set comprising 7,993 examples. Therefore, we also follow the same evaluation on this dev set to facilitate comparisons with their performance.
Secondly, to assess the model's \textbf{noise robustness and unknown robustness}, we extracted subsets from the above test sets that contained relevant documents in the retrieved list. We then enumerated each retrieved document to determine if it was a golden document for the given question. Based on the noise ratio $r$, for instance, if the top-$k$ documents are needed for the generator, then $k \cdot r$ would be the number of noisy documents, and $k \cdot (1-r)$ would be the number of relevant documents. For example, when noise ratio is 20\% and top-5 documents are needed, then 4 are relevant documents, and 1 is irrelevant documents. 
During the enumeration of the retrieved documents in data pre-processing, we populated two lists; when one list reached its limit, we stopped adding more documents to that list until both lists were complete. 
In instances where no relevant documents are retrieved by the DPR for certain questions, we exclude these from robustness evaluation. Therefore, the subset is smaller than the original test set, as shown in Table \ref{tab:stats}.

\begin{table}
\centering
\setlength{\tabcolsep}{2mm}{
\scalebox{0.9}{\begin{tabular}{l|ccr}
\toprule
\multirow{2}{*}{Models $\downarrow$} & \multicolumn{3}{c}{RealTimeQA}  \\
 & EM & F1 & RR \\
\midrule
Retrieve-Read~\cite{shi2023replug} & 15.6 & 19.9 & 6.1 \\
\ \ + \textsc{Chain-of-Note} (ours) & 15.7 & 20.3 & 13.0 \\
\bottomrule
\end{tabular}}}
\vspace{-0.1in}
\caption{Evaluation on Unknown Robustness. The \textsc{CoN} shows better performance than standard RALM system.}
\vspace{-0.1in}
\label{tab:rr}
\end{table}

\subsubsection{Baseline Methods}

\textsc{Chain-of-Note (CoN)} is built upon the traditional retrieve-then-read pipeline~\cite{lewis2020retrieval}. Recent implementations such as \citet{lazaridou2022internet,shi2023large,luo2023sail} integrate large language models to achieve better performance. Therefore, we primarily compare our approach against these retrieve-read methods.
As outlined in the $\S$\ref{sec:method}, we denote an input question as $x$ and its corresponding answer as $y$. Besides, $d_i$ represents the $i$-th retrieved document, and $y_{d_i}$ is the associated reading note for that document. Here we show the difference of methods to compare.

\vspace{0.05in}
\noindent \textbf{QA fine-tune w/o IR} are trained to directly generate an answer from the input question, without relying on any external retrieved information. Essentially, it learns the function $f: x \rightarrow y$, transforming the question $x$ directly to answer $y$.

\vspace{0.05in}
\noindent \textbf{Retrieve-Read~\cite{shi2023replug}} are trained to generate an answer not only from the question but also by incorporating retrieved documents. It learns the function $f: \{x, d_1, \cdots, d_k\} \rightarrow y$, meaning it transforms the question $x$ and a set of retrieved documents $\{d_1, \cdots, d_k\}$ into an answer $y$.

\vspace{0.05in}
\noindent \textbf{Retrieve-Read with \textsc{Chain-of-Note}} are trained to generate reading notes for each retrieved document before formulating the final answer. It learns the function $f: \{x, d_1, \cdots, d_k\} \rightarrow \{y_{d_1}, \cdots, y_{d_k}, y\}$, thereby enabling the model to process the question $x$ and retrieved documents $\{ d_1, \cdots, d_k \}$ to produce reading notes $\{ y_{d_1}, \cdots, y_{d_k} \}$ and the final answer $y$.

For fair comparability, we trained all LLaMa-2 models on same training set, with the main difference being in the input and output formats. 
\textbf{We also note that} the experiments conducted with GPT-4 were performed in a zero-shot setting. The prompts used for various experimental conditions are detailed in Appendix \ref{sec:prompt}.

\vspace{0.05in}
\subsubsection{Evaluation Metrics}
\label{sec:metric}

For the evaluation of open-domain QA performance, we have employed two widely recognized metrics: Exact Match (EM) and F1 score, as suggested by prior work in the \citet{chen2017reading,karpukhin2020dense,zhu2021retrieving}. 
For EM score, an answer is deemed correct if its normalized form -- obtained through the normalization procedure delineated by \cite{karpukhin2020dense} -- corresponds to any acceptable answer in the provided list.
Similar to EM score, F1 score treats the prediction and ground truth as bags of tokens, and compute the average overlap between the prediction and ground truth answer~\citep{chen2017reading}.
Besides, we use reject rate (RR) to evaluate the unknown robustness when given questions beyond a language model's knowledge scope.

Finally, since GPT-4 is not directly trained on open-domain QA benchmarks, employing EM / F1 for evaluation is challenging. Therefore, we adopt the approach outlined in \citet{mallen2023not,kandpal2023large}, utilizing accuracy as the evaluation metric. Accuracy considers a prediction correct if any substring of the prediction exactly matches any of the provided correct answers.

\subsection{Evaluation on Overall QA Performance}

Table \ref{tab:overall} demonstrates that the RALM consistently outperforms the directly fine-tuned LLaMa-2 with QA pairs, without retrieval. This improvement is closely tied to the effectiveness of the retrieval process. As indicated in Table \ref{tab:stats}, DPR demonstrates markedly superior retrieval performance on the NQ and TriviaQA datasets compared to WebQ. Consequently, the benefits of retrieval are more pronounced on NQ and TriviaQA.
Furthermore, when comparing our enhanced RALM, which integrates \textsc{CoN}, with the standard RALM, our method persistently shows better performance. There is an average improvement of +1.97 in EM scores across all three datasets when using LLaMa-2 as backbone language model. 
Delving deeper, we find that this improvement varies depending on whether DPR successfully retrieves relevant documents. Specifically, the average improvement is +1.2 when DPR retrieves relevant documents and +2.3 when it does not on the NQ dataset. This disparity suggests that our \textsc{CoN} improve RALM's in scenarios where more noisy documents are fetched in the first retrieval stage. This observation aligns with our findings on noise robustness, which are elaborated in the subsequent sections detailing our experimental results.

Furthermore, the dynamics observed with larger language models differ from those noted in experiments with smaller-sized models due to their superior factual knowledge. The impact of utilizing retrieval is observed to be less pronounced with larger models and can even be detrimental in certain cases, such as with TriviaQA, where questions are mostly straightforward. Concerning the comparison between \textsc{CoN} and the baseline, the performance trend remains consistent with that observed in smaller-sized models, suggesting that \textsc{CoN} maintains its significance across different model sizes.

\subsection{Evaluation on Noise Robustness}

As illustrated in Table \ref{tab:overall}, when faced with entirely noisy documents, both the standard RALM and our \textsc{Chain-of-Note} enhanced RALM underperformed compared to the no-retrieval setting. 
This suggests that RALMs can be misled by noisy information, leading to more hallucinations. 

Notably, equipping the model with \textsc{CoN} enables it to perform nearly as well as the baseline model directly fine-tuned with QA pairs without retrieval, showcasing its robustness to noise and its ability to disregard irrelevant information. The \textsc{CoN} approach is effective not only in fine-tuned, smaller-sized models but also in large language models, such as GPT-4, with adjustments made only to the prompt. Besides, in comparison to the \textsc{Chain-of-Thought} technique, commonly utilized in reasoning scenarios, \textsc{CoN} presents a more efficient strategy for retrieval-augmented settings, particularly in addressing knowledge-intensive tasks.

% Our RALM with \textsc{CoN} not only outperforms the standard RALM but also exceeds the performance of LLaMa-2 without IR, thus significantly enhancing RALM's noise robustness. Additionally, training the model to generate a series of notes helps balance the fine-tuning process, preventing overfitting and enhancing performance even in the absence of relevant documents.
% \vspace{0.05in}
% \noindent \textbf{Noisy Ratio.} Figure \ref{fig:noise-ratio} 
Table \ref{tab:noise} shows that RALM enhanced with \textsc{CoN} consistently outperforms the standard RALM, especially in scenarios with exclusively noisy documents. An average improvement of +7.9 in EM score on fully noisy documents is observed on three open-domain QA datasets, in average. Experiments with lower noise ratios also consistently demonstrate the improvements brought by \textsc{CoN}, aligning the overall QA performance.

% \vspace{0.05in}
% \noindent \textbf{Noisy Type.} Our evaluation of noise robustness was carried out under two scenarios: using noise documents obtained through retrieval (by removing relevant documents from the retrieved sets and retaining the top-ranked irrelevant ones) and using completely random documents sampled from the entire Wikipedia. Noisy retrieved documents often contain misleading information due to their semantic similarity to the input question, contrasting with random documents which represent total noise.
% Our comparison with random noise revealed several important observations. Figure \ref{fig:dpr-llm} illustrates that both standard RALM and RALM with \textsc{CoN} perform better with random documents than with noisy retrieved ones. This indicates that semantically relevant noisy documents are more likely to mislead the language model into producing incorrect information. Moreover, in both noisy scenarios, our method shows enhanced robustness compared to the standard RALM.

\subsection{Evaluation on Unknown Robustness}

Table \ref{tab:rr} illustrates that our RALM equipped with \textsc{CoN} exhibits superior robustness in handling unknown scenario, particularly evident in the RealTimeQA benchmark. This benchmark falls completely outside the model's domain and contains real-time information that was not part of the LLaMa-2 pre-training data. Despite this, models are still capable of providing correct answers in some cases, as the answers remain consistent over time. In comparison to the standard RALM system, our method shows a significant improvement, exceeding +10.5 in its ability to reject to answer questions in unknown scenario. The evaluation is based on reject rate (RR), i.e., number of rejected questions / total questions. This highlights our model's enhanced capability to discern and disregard information that is unfamiliar or not learned during its initial training phase.

\begin{figure}[t]
    \centering
    {\includegraphics[width=0.5\textwidth]{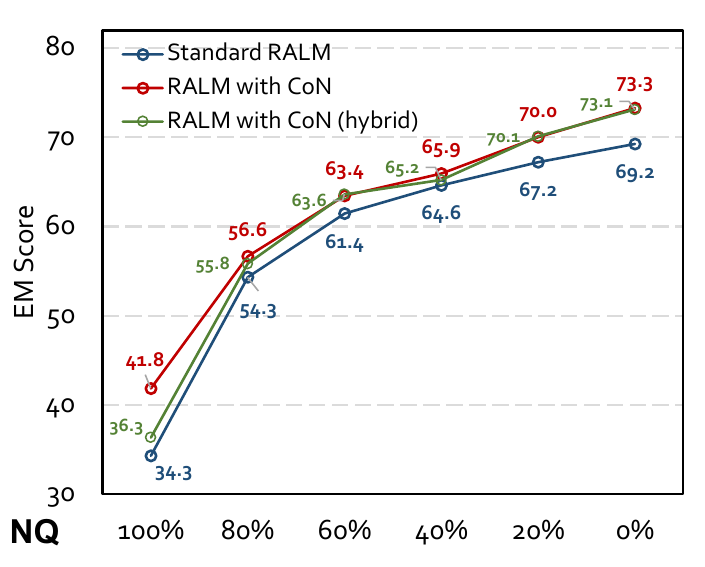}\label{fig:em-noise-nq}}
    \vspace{-0.3in}
    \caption{Using a hybrid training strategy demonstrates slightly lower robustness across various noise ratios but consistently better performance than standard RALMs.}
\label{fig:noise-ratio}
\end{figure}

\begin{table}
\centering
\setlength{\tabcolsep}{1.8mm}{
\scalebox{0.9}{\begin{tabular}{l|c}
\toprule
Models $\downarrow$ & Inference Time(s)  \\
\midrule
Retrieve-Read & \ \ 0.6104 \\
\ \ + \textsc{Chain-of-Note} & 12.0192 \\
\ \ + \textsc{Chain-of-Note} (hybrid) & \ \ 0.6074 \\
\bottomrule
\end{tabular}}}
\vspace{-0.1in}
\caption{The inference time comparison shows the average decoding time per example on 8$\times$A100 GPUs.}
\vspace{-0.1in}
\label{tab:time}
\end{table}

\subsection{Evaluation on Hybrid Training Strategy}
\label{sec:hybrid}

As illustrated in Figure \ref{fig:noise-ratio} and Table \ref{tab:time}, our proposed RALM equipped with a hybrid strategy demonstrates slightly lower robustness across various noise ratios while but keeping similar effcient decoding time consumption to the standard RALM. This indicates that our \textsc{Chain-of-Note} framework, when implemented with a hybrid training strategy, is highly applicable to a wide range of real-world business scenarios. This enhancement in robustness without significant time overhead highlights the practical value and efficiency of our approach, making it a viable solution for environments where QA accuracy can vary but inference time  is crucial.

%% file: 6-Related.tex
Retrieval-Augmented Language Models (RALMs) represent a significant advancement in natural language processing, combining the power of large language models with the specificity and detail provided by external knowledge sources~\citep{guu2020realm,lewis2020retrieval,izacard2022few}. Recent studies highlight the impact of context relevance on language model performance \citep{creswell2022selection,shi2023large,yoran2023making}. Notably, \citet{creswell2022selection} demonstrated that incorporating random or irrelevant contexts could adversely affect QA performance. In contrast, \citet{shi2023large} discovered that adding irrelevant context to exemplars or task-specific instructions can sometimes enhance model performance, implying that models might intrinsically possess capabilities, developed during pre-training, to manage such scenarios. Most pertinent to our research is the study by \citet{yoran2023making}, which focused on training RALMs to disregard irrelevant contexts. This approach, while distinct from our proposed solution, underscores the importance of context relevance in enhancing the effectiveness of RALMs.

Besides, we present more related Chain-of-\textbf{X}s approaches (e.g., Chain-of-Thought (CoT)~\citep{wei2022chain}) in the Appendix \ref{sec:r1} and \ref{sec:r2}.

%% file: 7-Conclusion.tex
In this paper, we introduce the \textsc{Chain-of-Note (CoN)} framework, a novel methodology designed to enhance the robustness of RALMs. The central concept of \textsc{CoN} revolves around the generation of sequential reading notes for each retrieved document. This process allows for an in-depth assessment of document relevance to the posed question and aids in synthesizing this information to craft the final answer. 
Our experiments show that GPT-4, when equipped with \textsc{CoN}, outperforms the \textsc{Chain-of-Thought} approach. 
Besides, we utilized GPT-4 to create 10K \textsc{CoN} data, subsequently trained on a LLaMa-2 7B model.
Our experiments across four open-domain QA benchmarks show that RALMs equipped with \textsc{CoN} significantly outperform standard fine-tuned RALMs.

%% file: 8-Limitation.tex
One major limitation of the \textsc{Chain-of-Note (CoN)} approach is its increased inference cost due to the sequential generation of notes. 
While \textsc{CoN} is beneficial for assessing the relevance and integrating external knowledge, it results in longer response times, which is problematic for time-sensitive applications. 
Moreover, the system's efficiency depends on the conciseness and relevance of the generated notes, which can fluctuate based on the complexity of the retrieved documents.

%% file: 9-Appendix.tex
\subsection{More Related Work}

\subsubsection{Retrieval-Augmented Language Models}
\label{sec:r1}
Retrieval-Augmented Language Models (RALMs) represent a significant advancement in natural language processing, combining the power of large language models with the specificity and detail provided by external knowledge sources~\citep{guu2020realm,lewis2020retrieval,izacard2022few}. These models first leverage a retriever to scan a vast evidence corpus, such as Wikipedia, to identify a set of documents pertinent to the user's query. Following this, a reader component is employed to meticulously analyze these documents and formulate a response. This two-pronged approach ensures both relevance and depth in the generated answers.
Recent follow-up work has mainly focused on improving the retriever~\citep{karpukhin2020dense,qu2021rocketqa,sachan2022questions,ma2023chain} or the reader~\citep{izacard2021leveraging, cheng2021unitedqa,yu2022kg}, training the system end-to-end~\citep{lewis2020retrieval,singh2021end}, and integrating the retrieval systems with large-scale black-box language models~\citep{yu2022generate,shi2023replug,yu2023improving,trivedi2023interleaving}.
Another line of RALMs such as kNN-LM~\citep{khandelwal2020generalization,zhong2022training} retrieves a set of tokens and interpolates between the next token distribution and kNN distributions computed from the retrieved tokens at inference.
The evolution has also led to the emergence and popularity of retrieval-augmented products, such as ChatGPT plugin, Langchain, and New Bing.

\subsection{Chain-of-\textbf{X} Approaches in Large Language Models}
\label{sec:r2}
Recent research shows that large language models (LLMs) are capable of decomposing complex problems into a series of intermediate steps, pioneered by the concept of Chain-of-Thought (CoT) prompting~\citep{wei2022chain,kojima2022large}. 
The CoT approach mirrors human problem-solving methods, where complex issues are broken down into smaller components. By doing so, LLMs can tackle each segment of a problem with focused attention, reducing the likelihood of overlooking critical details or making erroneous assumptions. This sequential breakdown makes the reasoning process more transparent, allowing for easier identification and correction of any logical missteps. 

The CoT methodology has been effectively applied in various contexts, including multi-modal reasoning~\citep{zhang2023multimodal}, multi-lingual scenarios~\citep{shi2022language}, and knowledge-driven applications~\citep{wang2023knowledge}. 
And additionally, there has been a surge in the development of other chain-of-X methods, addressing diverse challenges in LLM applications. These include  chain-of-explanation~\citep{huang2023chain}, chain-of-knowledge~\citep{wang2023boosting}, chain-of-verification~\citep{dhuliawala2023chain} and IR chain-of-thought~\citep{trivedi2023interleaving}.
For instance, Chain-of-Verification~\citep{dhuliawala2023chain} generates an initial response, formulates verification questions, and revises the response based on these questions, reducing factual errors and hallucinations in the response. Closely related to our work is IR chain-of-thought~\citep{trivedi2023interleaving}, which employs CoT to infer and supplement unretrieved information, thereby improving the accuracy of complex reasoning tasks.
While chain-of-\textbf{X} approaches have shown promise in enhancing LLMs' performance across various domains, their application in RALMs, particularly for improving robustness in noisy and unknown scenarios, is relatively unexplored. This gap signifies further research in applying these strategies to augment RALMs, thereby enhancing their robustness and reliability.

% Besides, Chain-of-Knowledge~\citep{wang2023boosting} elicits LLMs to generate explicit pieces of knowledge evidence in the form of structured triple. This is inspired by human behaviors that we often draw a mind map or knowledge map as the reasoning evidence in the brain before answering a complex question.
% Similarly, the Chain-of-Knowledge method proposed by Wang et al.\cite{wang2023boosting} encourages LLMs to produce explicit knowledge evidence in structured formats, akin to creating mental or knowledge maps.

\subsection{Dataset Information}
\label{sec:dataset}

-- TriviaQA~\citep{joshi2017triviaqa} contains a set of trivia questions with answers originally scraped from trivia and quiz-league websites.

\noindent -- WebQ~\citep{berant2013semantic} consists of questions selected using Google Suggest API, where the answers are entities in Freebase.

\noindent -- NQ~\citep{kwiatkowski2019natural} were collected from real Google search queries and the answers are one or multiple spans in Wikipedia articles identified by human annotators. 

\subsection{Implementation Details}

In the retrieval phase, we employed DPR~\cite{karpukhin2020dense} to retrieve documents from Wikipedia. 
We accessed the model via direct loading from the official DPR repository hosted on GitHub.
Subsequent to retrieval, our fine-tuning process for the LLaMA-2~\cite{touvron2023llama} model runs for 3 epochs with a batch size set to 128, leveraging the DeepSpeed library \cite{rasley2020deepspeed} and the ZeRO optimizer \cite{ma2021zero}, with bfloat16 precision.
The learning rates are set to \{$1e$-$6$, $2e$-$6$, $5e$-$6$, $1e$-$5$, $2e$-$5$\}, and the empirical results indicated that 
$5e$-$6$ yielded the best model performance, hence we standardized the learning rate for all reported numbers.
Greedy decoding is applied during inference on all experiments to ensure deterministic generations.

\subsection{Instruction Prompts}
\label{sec:prompt}

\noindent (1) For standard RALM, the instruction is: 

\vspace{0.05in}
\noindent Task Description: The primary objective is to briefly answer the question.
\vspace{0.15in}

\noindent (2) For RALM with \textsc{CoT}, the instruction is: \

\vspace{0.05in}
\noindent Task Description: Think step by step to answer the question.
\vspace{0.15in}

\noindent (3) For RALM with \textsc{CoN}, the instruction is: \

\vspace{0.05in}
\noindent Task Description: \\
1. Read the given question and five Wikipedia passages to gather relevant information. \\
2. Write reading notes summarizing the key points from these passages. \\
3. Discuss the relevance of the given question and Wikipedia passages. \\
4. If some passages are relevant to the given question, provide a brief answer based on the passages. \\
5. If no passage is relevant, direcly provide answer without considering the passages. \\

\begin{table*}
\centering
\setlength{\tabcolsep}{1.5mm}{\scalebox{0.92}{
\begin{tabular}{p{16.5cm}}
\toprule
\textbf{Question:} When was the last Summer Olympics in the USA \ \ \textbf{Answer:} 1996 \\
\textbf{Wikipedia document \#1:} In \textcolor{red}{Chicago's 2016 Summer Olympics bid}, the games would have been held from July 22 to August 7, with the Paralympics held between August 12 and 28. The plan emphasized use of Chicago Park District ... (Wikipedia page: Chicago bid for the 2016 Summer Olympics) \\
\textbf{Wikipedia document \#2:} \textcolor{green}{The 1996 Olympics are the most recent edition} of the Summer Olympics to be held in the United States. Los Angeles will host the 2028 Summer Olympics, 32 years after the games were held in Atlanta. ... (Wikipedia page: 1996 Summer Olympics) \\
\textcolor{gray}{(Other retrieved Wikipedia documents are omitted.)} \\
\textbf{Standard RALM:} \textcolor{red}{2016 \ \xmark} \\
\textbf{RALM with \textsc{Chain-of-Note}:} Wikipedia passage \#1 discusses Chicago's bid for the 2016 Summer Olympics, but does not mention whether it is a successful bid.
Wikipedia passage \#2 confirms that the 1996 Summer Olympics were the most recent edition of the Summer Olympics to be held in the United States. Based on the information given in above passages, the answer is \textcolor{green}{1996 \ \cmark} \\
\midrule
\textbf{Question:} The first Jnanpith Award recipient was an author in which language? \ \ \textbf{Answer:} Malayalam \\
\textbf{Wikipedia document \#1:} The first Jnanpith Award was given to G. Sankara Kurup in 1965 for his collection of poems, Odakkuzhal (\"The Bamboo Flute\"), ... (Wikipedia page: Jnanpith Award) \\
\textbf{Wikipedia document \#2:} ... The final round had four authors; Kazi Nazrul Islam (Bengali), D. V. Gundappa (Kannada), Viswanatha Satyanarayana (Telugu), and \textcolor{green}{G. Sankara Kurup (Malayalam)}. On 19 November 1966, Kurup presented with...  (Wikipedia page: Jnanpith Award) \\
\textcolor{gray}{(Other retrieved Wikipedia documents are omitted.)} \\
\textbf{Standard RALM:} \textcolor{red}{G. Sankara Kurup \xmark} \\
\textbf{RALM with \textsc{Chain-of-Note}:} Wikipedia passage \#1 states that the first Jnanpith Award was given to G. Sankara Kurup in 1965 for his collection of poems, Odakkuzhal. Wikipedia passage \#2 mentions that the Jnanpith Award was given to G. Sankara Kurup in 1966 for his work in Malayalam. Based on the information given in above passages, the answer is \textcolor{green}{Malayalam \cmark} \\
\bottomrule
\end{tabular}}}
\vspace{-0.1in}
\caption{Case Study. Compared to Standard RALM, our RALM with \textsc{Chain-of-Note} exhibits a deeper understanding of how documents reveal information relevant to the question. It goes beyond merely capturing surface-level terms, leading to more accurate responses.}
\label{tab:case}
\end{table*}

\subsection{Case Studies}

In our case studies, as illustrated in Table \ref{tab:case}, we compare the responses generated by the standard RALM and our enhanced RALM with \textsc{CoT}. These examples highlight the differences in how each model processes and interprets information from retrieved documents.

The first case shows a question pertains to the most recent Summer Olympics held in the USA. The standard RALM is misled by the mention of "Chicago's bid for the 2016 Summer Olympics." Lacking a deep comprehension of the content, it incorrectly focuses on the more recent year (2016), resulting in an inaccurate answer. In contrast, the RALM with \textsc{CoN} carefully analyzes the information. It notes that while Chicago bid for the 2016 Olympics, there's no confirmation of it being a successful bid. This leads to the correct conclusion that the most recent Olympics in the USA were held in 1996.
The second case study involves identifying the language of the first Jnanpith Award recipient. Here, the standard RALM fails to synthesize information across documents. It identifies G. Sankara Kurup as the award recipient but does not connect this information to the language of his work. Conversely, the RALM with \textsc{CoN} effectively combines details from both documents. It recognizes that while the first document mentions Kurup's award, the second document provides the missing language detail, leading to the correct answer of Malayalam.

\subsection{Licenses}

The four open-domain QA benchmarks, LLaMa-2 models are all released under MIT License. They are all for research purposes, and our experiments are consistent with their intended usage.